%% file: RandDecisionTrees_arxiv.tex
\documentclass[10pt,english]{article}
\usepackage[utf8]{inputenc}
\usepackage{amsmath}
\usepackage{amsfonts}
\usepackage{amssymb}
\usepackage{graphicx}
\usepackage{color}
\usepackage{enumitem}
\usepackage{algorithm}
\usepackage{algorithmicx}
\usepackage{algpseudocode}
\usepackage{caption}
\usepackage{subcaption}
\usepackage{url}
\usepackage[left=2cm,right=2cm,top=2cm,bottom=2cm]{geometry}

\usepackage{palatino}

\newtheorem{definition}{Definition}
\newtheorem{theorem}{Theorem}

\newcommand{\RR}{\mathbb{R}}
\newcommand{\NN}{\mathbb{N}}

\newcommand{\norm}[1]{\ensuremath{\left\|#1\right\|_2}}

\newcommand{\ignore}[1]{}

\newcommand{\mat}[1]{ {\ensuremath{\mathsf{#1} }}}
\newcommand{\matA}{\mat{A}}
\newcommand{\matU}{\mat{U}}
\newcommand{\matV}{\mat{V}}
\newcommand{\matX}{\mat{X}}
\newcommand{\matSig}{\mat{\Sigma}}

\newcommand{\y}{\mathbf{y}}

\title{Non-uniform Feature Sampling for Decision Tree Ensembles}

\author{
  Anastasios Kyrillidis\\
  Computer and Communication Sciences, EPFL \\
  \texttt{anastasios.kyrillidis@epfl.ch}
  \and
  Anastasios Zouzias\thanks{The research leading to these results has received funding from the European Research Council under the European Union's Seventh Framework Programme (FP7/2007-2013) / ERC grant agreement $n^{o}$ 259569.}\\
   IBM Research Lab, Zurich \\
  \texttt{azo@zurich.ibm.com}
}

\begin{document}
\maketitle
\input{abstract}
\vspace{-0.05in}
\input{intro}
\vspace{-0.05in}
\input{related}
\vspace{-0.05in}
\input{problemSetting}
\vspace{-0.05in}
\input{feature}
\vspace{-0.05in}
\input{experiments}
\vspace{-0.05in}
\input{discussion}

\bibliographystyle{acm}
\bibliography{RandDecisionTrees}

\end{document}

%% file: abstract.tex
\begin{abstract}
We study the effectiveness of non-uniform randomized feature selection in decision tree classification. We experimentally evaluate two feature selection methodologies, based on information extracted from the provided dataset: $(i)$ \emph{leverage scores-based} and $(ii)$ \emph{norm-based} feature selection. Experimental evaluation of the proposed feature selection techniques indicate that such approaches might be more effective compared to naive uniform feature selection and moreover having comparable performance to the random forest algorithm~\cite{breiman2001random}.

% We experimentally show that non-uniform sampling might influence the computational complexity of constructing the classifier as well as its classification accuracy. %This phenomenon indicates that such more sophisticated selection process might be an approach that any machine learning practitioner should have under consideration.

%\textbf{The efficiency of classification algorithms is highly correlated with several parameters: $(i)$ the quality and quantity of the test data, $(ii)$ the equitable selection of problem parameters and, $(iii)$ the proper selection of features. In this work, we mainly focus on the affect that the feature selection has in classification tasks and how {\it biased} randomized procedures influence the complexity as well as the classification accuracy of the classifier. We propose two feature selection probability distributions based on information extracted form the training data using standard linear algebra techniques. To strengthen our proposal, we conduct experiments on real problem datasets to demonstrate the improvements in accuracy of our approach.}
\end{abstract}

%% file: intro.tex
\section{Introduction}
%\textcolor{red}{Big Data:}
Living in the era of Big Data, massive amount of information is now publicly available, aggrandizing our expectations for new developments, both in well-established and contemporary scientific tasks. 
However, this ever increasing data often contradicts with the principle of {\it parsimony}: in a high-dimensional feature space the proper selection of features, that results in succinct descriptions of the problem, cannot be easily derived. This fact jeopardizing the interpretability of the solution.
This curse of dimensionality can also pose difficulties with respect to the qualitative performance of methods, imperilling their accuracy as well as their robustness in the case of noise and outlier presence.
%this ever increasing amount of data poses a compulsory obstacle to overcome: that of {\it high-dimensional computing}.
%Moreover, large scale problems stretch the capabilities of algorithms to their limits, due to the intensive computational requirements. 
%Furthermore, high-dimensional data analysis has deeply changed the foundations of traditional statistics: new theoretical cornerstones are needed to support the known-until-now practices. % within the Big Data area. 

%\textcolor{red}{Important application where Big data applies is classification:} 
An important application that suffers from this difficulty is classification. 
An abstract description for the case of binary classification is given below:
\vskip.1in
\noindent \textsc{Binary Classification Problem:} {\it Assume \vskip-.2in
\begin{align}\nonumber
\mathcal{D}_{\texttt{train}} = \left\{\left(\matX_1, y_1\right), ... \left(\matX_n, y_n\right)\right\} 
\end{align} \vskip-.1in 
\noindent be a collection of $n$ supervised train feature vectors $\matX_i \in \mathbb{R}^d$ with corresponding labels $y_i\in \left\{\pm 1\right\}$. Given $\mathcal{D}_{\texttt{train}}$, we want to learn a classifier $\mathcal{C}: \mathbb{R}^d \rightarrow \lbrace \pm 1\rbrace$ such that, for an \emph{unsupervised} input $\mathcal{D}_{\texttt{test}} = \left\{\matX_j : \matX_j \notin \mathcal{D}_{\texttt{train}}\right\}$, $\mathcal{C}$ computes labels on the elements of $\mathcal{D}_{\texttt{test}}$ with the lowest possible classification error.}
\vskip.1in
\noindent Several cases have been reported in the literature where classification using the over-complete set of features (i.e., without proper selection or pre-processing) can be as poor as random guessing, due to noise accumulation in the high-dimensional feature space \cite{fan2008high}.\footnote{The authors in \cite{fan2008high} demonstrate further that almost all linear discriminants can perform as poorly as the random guessing.} 
Wherefore, irrelevant or redundant information ``interfere'' with useful one and its removal could gain in classification. 
Fortunately, common wisdom indicates that, in practice only a few features are important for classification and thus such removal is applicable \cite{amit1997shape}; e.g., in DNA data \cite{west2001predicting, fan2006statistical}, only a few genes are influential in a gene sequence expression. 
Moreover, an excessive number of attributes usually results in prohibitive running times and storage requirements during training for real-time applications; in memory-limited cases, further post-processing is required \cite{kulkarni2012pruning}.
%Moreover, utilizing excessive number of features might lead to classifiers with higher construction complexity, where either more computational power and storage is needed to accommodate the complex structures or pruning techniques need to be applied.

\ignore{
From a different perspective, the current trend in data mining assumes a {\it streaming model} \cite{abdulsalam2011classification, mitliagkas2013memory, aggarwal2004demand}: only a snapshot of data is available per session, due to high storage complexity otherwise. %, often arriving at high rates.
Thus, the number of available samples $n$ might be inadequate, as compared to the ambient dimension $d$ of the feature set $\matX_i, ~\forall i$, which weakens generalization properties of the classifier. 
%Furthermore, 
%To accommodate such system dynamics, solvers have to work on-the-fly, extracting information only from a small snapshot of the data, as compared to $d$.
}
In stark contrast, training a well-behaved {\it individual} classifier with a {\it predetermined and fixed subset} of features over a restricted train dataset is a difficult task; it often creates overfitting issues, where the loss of generalization is observed on incoming new data. 
To overcome this difficulty, recent developments \cite{ho1998random, ho1995random, hansen1990neural}, based on \cite{kleinberg1990stochastic}, have proposed the systematic construction of classifiers:
randomly and independently selected subsets of features are used per learner and the final decision is taken as a {\it majority (averaging) rule} over the collection of learners for the given data; such structures are generally known as classifier ensembles \cite{dasarathy1979composite}. 

However, a {\it naive selection} of features might still doubt the practicality of classifier ensembles in such settings: the random selection might lead to extremely ``weak'' learners, increasing drastically the number of ensemble components required for a desired classification error. 
Moreover, the authors of \cite{joly2012l1} highlight the exponentially increasing space-complexity of tree-based ensembles to achieve a given accuracy; thus, more sophisticated selection procedures might lead to less expensive constructions.
%  Number of nodes of a tree grows as $nM$ where $n$ is the size of the learning sample and $M$ is the number of trees in the ensemble. For increased number of features, we need more trees for a given accuracy. Thus, the space complexity of the tree ensembles grows as $nM(p)$ where $p$ is the feature complexity, which doubts the practicality in large scale problems or when the memory is limited.\footnote{As an alternative way to reduce complexity of the learning ensemble, recent works have proposed pruning of the trees.}

To this end, a compromise between these two extremes is imperative in practice: by judiciously selecting a subset of {\it significant} attributes, allowing randomness during the selection, one can achieve acceptable classification accuracy and desired generalization attributes with low space complexity.

\vskip.1in
\noindent \textbf{Our contributions:} 
In this context, sophisticated dimensionality reduction techniques might play a crucial role. Instead of selecting features uniformly at random, we utilize linear algebraic techniques to ``bias'' the selection procedure. Based on the work \cite{drineas2008relative, mahoney2012fast, rudelson2007sampling}, we use matrix-based information scores to define a non-uniform probability distribution that favors more dominant features. %The proposed classifier operates in a {\it dynamic} fashion, where both training and test samples are obtained on-the-fly and the weighting scheme is updated systematically. \vskip-.1in
%\item [$(ii)$] We use low-dimensional projections using random vectors to ``mix'' the features in a low dimensional subspace.  \vskip-.1in
%\end{itemize}
%Thus, it is important to select a subset of important features for classification. 
%While there is only a fraction of features that account for most of the variation in the data such as tumor classification using gene expression data, using all features will increase the misclassification rate.
%Since computing... we assume a streaming model to use incremental SVD.
%\textcolor{red}{
%$(ii)$ mention that ideas from compressed sensing and randomized algebra can be used to linearly combine the features in a low dimensional subspace. Do these projections find low subspaces where the mixing of variables result in small classification errors? $(iii)$ Experiemental results and what we observe highlights. Mention that in this work, we focus on tree-based ensemble methods as a classification technique, based on decision tree structures.}

Empirical results show an overall improved classification capability using our approach, as compared to classic state-of-the-art schemes for a given training time period. 

%% file: related.tex
\section{Related Work}
As already mentioned above, a classical technique in classification focuses on the idea of {\it ensemble classifiers}: by combining a set of ``weak'' learners that approximate the training data, one could obtain a ``strong'' classifier, i.e., a classifier with better generalization performance (see e.g.,~\cite{adaboost}). Based on this approach, one can generate several ``weak'' learners by applying one or combination of the following designs: \vskip-.1in
\begin{itemize}[leftmargin=.2in]
\item [$(i)$] \emph{Feature selection}: each classifier is trained over a selected subset of features---e.g., for an excellent introduction, see the subspace method proposed in \cite{ho1998random, ho1995random} and the Randomized C4.5 algorithm in \cite{dietterich2000ensemble}, following the work \cite{amit1997shape}. \vskip-.1in
\item [$(ii)$] \emph{Training data subsampling and reweighting:} each classifier is trained over a subset of the training samples; then, the sample selection scheme is re-weighted, based on the classification error in the previous iteration---e.g., see the celebrated Boosting technique \cite{schapire2003boosting} and references therein.\footnote{To use such approach, one assumes many passes over the data.} \vskip-.1in
\item [$(iii)$] \emph{Linear combination of features under random low-dimensional embeddings,} where each learner uses the whole spectrum of features, trasformed by random linear mappings. %This guarantees that each ``weak'' classifier uses all the features in a compressed domain. \vskip-.1in
\end{itemize} 
There are several works in the literature where combinations of the above designs are used in practice---e.g., Random Forests with feature selection and data subsampling \cite{breiman2001random}. 

In \cite{amit1997shape, ho1995random, dietterich2000ensemble, fan2003random}, the authors consider case $(i)$ where the {\it random} model is proposed: to train a ``weak'' learner, each feature is selected uniformly at random, ignoring any prior information. 
While such strategy is maximally ``unbiased'' and easy to implement, it might lead to lower classification accuracy when a small number of features is selected each time or to higher complexity, due to the larger amount of classifiers required for a given accuracy level.
In \cite{ho1998random, fan2003random}, the authors further extend this strategy to {\it node optimization} for tree-based ensembles: at every level of each decision tree in the ensemble, the splitting decision rule over each node is derived using a random subset of features; in this work, our proposal does not consider this case and we leave it as a future research direction.

In this context, \cite{breiman2001random, fan2003random} show that randomization yields strict improvements over simple deterministic selection heuristics.\footnote{A non-exhaustive list of such rules in the case of tree-based classifiers includes Gain ratio and Gini index, designed to result into simple models for classification.} 
While simplicity helps, randomization is particularly useful in ``adversarial'' settings where ``bad'' features are present. \cite{breiman2001random} states the accuracy of RFs is insensitive to randomness in practice, see also~\cite{fan2003random}. 
However, subsequent developments on tree-based classifiers do not espouse this virtue. 
The author in \cite{robnik2004improving} proposes the \texttt{ReliefF} feature evaluation metric: the significance of each attribute is inversely proportional to what extent its values separate similar observations into different classes;
%The total computational complexity of \texttt{ReliefF} is $\mathcal{O}(n \log (n) d)$; 
details of this algorithm are provided in \cite{robnik2003theoretical}. 
\cite{breitenbach2003probabilistic} proposes the \emph{Probabilistic RFs} where feature selection procedure is further linearly transformed with linear kernels.
Such works amplify the suspicion that one can achieve more by applying sophisticated selection strategies rather than the blind randomization model.

In this paper, we attempt to raise this suspicion even more: we study several matrix-based sampling measures in order to signify important features in the selection procedure for higher classification accuracy. 

%Given a trained classifier, an important ingredient in the classification modeling is data mining. 
%Traditional approaches in classification usually nullify time coordinate from the learning process: train data is parsed ignoring the time-coherence between samples. 
%Nevertheless, in several applications, both labeled and unlabeled data are created on-the-fly, resulting in a system that evolves over time \cite{aggarwal2004demand}.
%As a concrete example, one can consider web-based applications where subset of products/movies/services are classified by the clients and can be used for training, while others remain unclassified; by taking into account the time stamp of every input, one can implicitly incorporate current {\it trends} in the classification procedure.

%% file: problemSetting.tex
\section{Problem setting}
\begin{figure}[tp]
\centering
\includegraphics[width=0.4\linewidth]{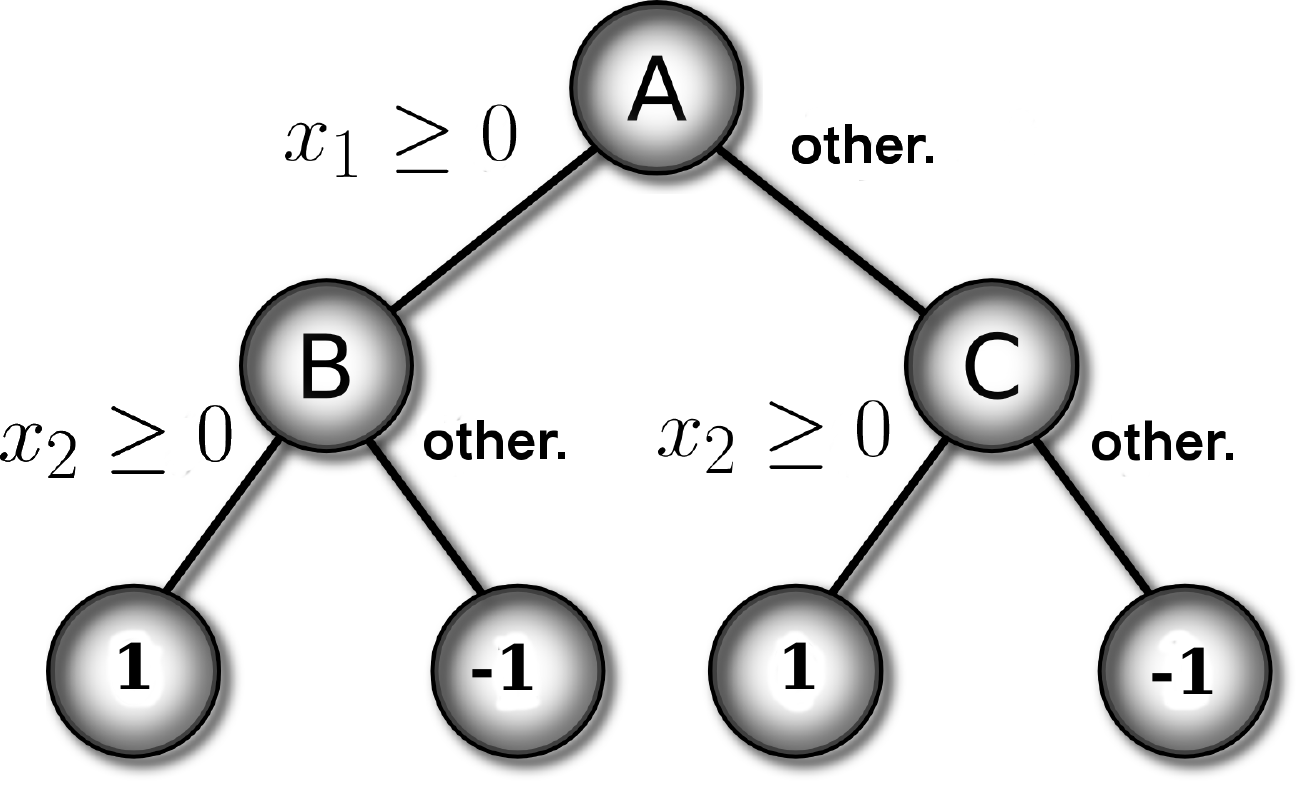}
\caption{Toy-example decision tree with two features $x_1, x_2$. Here, $A$ denotes the full sample set, $B, C \subseteq A$ are subsets satisfying the decision rule on node $A$. Leaf nodes contain only samples from class $1$ or $-1$.}
\label{fig:dec_tree}
\end{figure}

Throughout this paper, we use {\it decision trees} as ``weak'' learners; an illustrative example for a binary decision tree is given in Figure~\ref{fig:dec_tree}. 
These structures have inherent interpretation capabilities due to the explicit decision rules on the splitting nodes, are non-parametric, easy to implement and extremely fast to train, as compared to other classifiers. A non-exhaustive list of alternatives include linear classifiers \cite{witten2011penalized, liu2002gabor}, Support Vector machines \cite{cortes1995support, gunn1998support}, Neural Networks \cite{zhang2000neural}, etc; we conjecture that our proposed feature selection scheme can be easily applied to these cases and we leave this research direction for future work. 
%

%
%However, a single decision tree, either binary or multi-way, tends to over-fitting and have poor generalization performance with high variance and bias \cite{trevor2001elements}.
%%%%%%%%%%%%%%%%%%%%%%%%%%%%%%%%%%%%%%%%%%%%%%%%%%%%%%%%%%%%%%
%\vspace{-\baselineskip}
In the realm of random decision tree ensembles, we generate a set of decision trees, built on a subset of the initial feature set. We construct the ensembles as follows: for each tree, we sample non-uniformly and independently at random a set of $k$ features. 
Then, we train a decision tree classifier in its entirety (with a deterministic splitting criterion), restricted on these $k$ features.

It is important to notice that the above randomized procedure is quite different than Breiman's random forest algorithm (RFs)~\cite{breiman2001random, ho1998random}. In a parameter free implementation of RFs without bagging, each tree of the forest utilizes randomness in the splitting process of its construction. Conventional wisdom indicates that, at each node during the tree growing process, an uniformly random (and possibly different) sample of $\sqrt{d}$ features is utilized.
%

%Since independent realizations of classifiers on the training data generate individual errors, the majority of those with the correct classification rules will dominate over the defective ones (according to the majority vote rule). To this end, RFs have significantly higher accuracy and less over-fitting as compared to individual decision trees. 

\section{Our approach in a nutshell}
%Matrices provide a natural structure for encoding information. 
To describe the main ideas of our approach, assume that we represent the training dataset of $n$ objects with $d$ features as an $n\times d$ real\footnote{We implicitly assume that features are real-valued.} matrix $\matA$.
\ignore{
, i.e., 
\begin{align}
\matA = \begin{bmatrix}
\rule[.5ex]{3.5em}{0.4pt} & \matX_1^T & \rule[.5ex]{3.5em}{0.4pt} \\
\rule[.5ex]{3.5em}{0.4pt} & \matX_2^T & \rule[.5ex]{3.5em}{0.4pt}\\
 & \vdots & \\
\rule[.5ex]{3.5em}{0.4pt} & \matX_n^T & \rule[.5ex]{3.5em}{0.4pt}\\
\end{bmatrix}_{n \times d}.
\end{align}}
We propose the \texttt{LEverage ScoreS (LESS)} tree ensemble algorithm, a two-phase classifier construction, as reported in Algorithm \ref{algo:lev}. Let $\Pi := \left\{ \pi_1, \pi_2, \dots, \pi_d \right\}$ denote a probability distribution over the set of features that signifies the importance of features over $\mathcal{D}_{\texttt{train}}$. In the first phase, we compute $\Pi$, based on ideas described in Section \ref{sec:feature}. Next, we ``feed'' $\Pi$ into the second phase of our approach where: $(i)$ we select $k$ features according to $\Pi$ and, $(ii)$ based on these features, we generate $t$ decision trees.
\begin{algorithm}{}
	\caption{\texttt{LEverage ScoreS (LESS) Trees}}\label{algo:lev}
\begin{algorithmic}[1]
\Procedure{\texttt{LESS}}{$\matA$, $\y$, $t$, $k$} \Comment{$\matA\in\RR^{n\times d}, \y\in{\{\pm 1\}}^n$} \\
\Comment{$t, k\in \NN$: \# of trees and features}%\\
%\Comment{$k$: \# of features}~~~~~~~~~~~~~~
\State Compute $\Pi$, according to Eqn.~\eqref{eq:levscore}.
\For {$k=0,1,2,\ldots, t - 1 $ }
	\State Sample $k$ features of $\matA$ using $\Pi$.
	\State Construct $\matA^{(k)}\in \RR^{n\times k}$ restricted to the $k$ features. 
	\State Train decision tree using $(\matA^{(k)}, \y)$
\EndFor
\State Output: collection of $t$ trees.
\EndProcedure
\end{algorithmic}
\end{algorithm}
Finally, the collection of these decision trees is gathered and a standard majority voting scheme is applied to derive the predicted labelling of the model. Namely, given a set of decision trees and an unlabelled example, the algorithm returns as its predicted label the most frequent label over all the decision trees. 

Several remarks can be highlighted about the above algorithm. First, each decision tree is constructed on only a subset of features of size $k$ (usually $k$ is between $10$ and $50$); hence, as we show in Section \ref{sec:experiments}, Algorithm~\ref{algo:lev} has computational advantages over models that compute the best split over the whole feature set. 
Along the same lines, the resulting collection of trees are more interpretable than RFs since each tree depends only on a small set of features. Last, the random process described in Algorithm~\ref{algo:lev} is simpler than the random forest algorithm~\cite{breiman2001random} and hence might be amenable to theoretical justification in the future.

%% file: feature.tex
\section{Feature Selection Schemes}{\label{sec:feature}}

%Apart from generalization issues using irrelevant features, 
Exploiting the full spectrum of features creates a tradeoff between interpretability and predictive accuracy.
%e.g., tree ensembles, trained over many features, lead to extensive and diverse decision rules which is usually much less interpretable from a single CART decision tree trained over the most significant set of features. 
Thus, an important step to process such large-scale data is to construct 
%a compressed representation of $\matA$ for better interpretation in light of a corpus of field-specific knowledge. 
%To select features, we need 
an ``importance score" for each column of $\matA$ to denote the influence of the corresponding feature. 
Given such measure, we can then sample a predefined number of features $k$ for each decision tree, based on these scores.

 In this section, we describe three subsampling techniques for feature selection: $(i)$ uniform sampling, $(ii)$ column squared-norm based sampling and, $(iii)$ leverage scores-based sampling~\cite{drineas2008relative} (to be defined shortly).
\vskip.05in
\noindent \textbf{Uniform sampling:} each feature is selected with equal probability. Both strategies, where features are selected with or without replacement, have been tested; cf., \cite{ho1995random, ho1998random}. We use this policy as the baseline performance in our experiments.
\vskip.05in
\noindent \textbf{Norm-based sampling:} Recent developments on geometric functional analysis have dictated that squared norm subsampling can approximate well large datasets incurring small spectral norm~\cite{rudelson2007sampling}. Namely, in our setting sampling the i-th feature with probability proportional to $\norm{\matA_i}^2$ where $\matA_i$ is the i-th column of $\matA$ and $\norm{\cdot}$ denote the Euclidean norm.
\ignore{
spectral norm:
%have shown how to approximate large data matrix from its submatrix, selected according to the following theorem.
\vskip-.07in
\begin{theorem}[\cite{rudelson2007sampling}]
Let $\matA$ have stable rank $r = \frac{\|\matA\|_F^2}{\|\matA\|_2^2} \leq \text{rank}(\matA) \leq n$ for $n \leq d$. Let $\epsilon, \delta	\in (0, 1)$ and $\xi$ be an integer such that
\begin{align}
\xi \geq C\left(\frac{r}{\epsilon^{4}\delta}\right) \log \left(\frac{r}{\epsilon^{4}\delta} \right).
\end{align} Then, sampling $\xi$ columns from $\matA$ \emph{with probability proportional to their normalized $\ell_2$-norm}, results into submatrix $\widehat{\matA} \in \mathbb{R}^{n \times \xi}$ where:
\begin{align}
\|\matA - \matA P_k\|_2 \leq \sigma_{k+1}(\matA) + \epsilon \|\matA\|_2
\end{align} where $P_k$ is the projection onto the top $k$ right singular vectors of $\widehat{\matA}$ and $\sigma_{k+1}(\matA)$ is the $(k+1)$-th singular value of $\matA$.
\end{theorem} Here, each $\pi_j \in \Pi$ is computed as a function of the $\ell_2$-norm of the columns of $\matA$, i.e., $\pi_j \propto \frac{\|\matA_{:, j}\|_2^2}{\|\matA\|_F^2}$. 
}
\vskip.05in
\noindent \textbf{Statistical leverage scores sampling:}
The goal of statistical leverage scores is to construct a judiciously-chosen nonuniform importance sampling distribution over the set of columns, based on factorizations of $\matA$, according to the following definition:
\vskip-.1in
\begin{definition}[Statistical leverage scores \cite{lev:old,drineas2008relative}]
Let $\matA  \in \mathbb{R}^{n \times d}$ be a data matrix with $n$ objects and $d$ features with $r := \text{rank}(\matA) \leq n$ for $n \ll d$. Moreover, let $\matA = \matU \matSig \matV^\top$ be its \emph{Singular Value Decomposition} (SVD) where $\matV \in \mathbb{R}^{n \times r}$ contains the set of right singular column vectors. The normalized statistical leverage scores over the set of columns of $\matA$ are defined as:
\begin{align}{\label{eq:levscore}}
\pi_j = \frac{1}{r} \sum_{i = 1}^r (v_i(j))^2, ~~\text{for}~j = 1, \dots, d,
\end{align} where $v_i(j)$ denotes the $j$-th entry of the $i$-th right singular vector.
\end{definition}
\vskip-.1in
We highlight that, while sampling schemes $(ii)$ and $(iii)$ are slower than the simple uniform sampling, both result into generally higher classification accuracy for given sampling volume as we show in Section \ref{sec:experiments}. Moreover, strategy $(iii)$ can be well-approximated using fast randomized algorithms \cite{mahoney2012fast}. We note that leverage scores converge to uniform sampling when the coherence of the data matrix is small and scheme $(i)$ turns to be the optimal; thus, scheme $(iii)$ can be considered as a more generic selection strategy that includes $(i)$ as a special case.

%% file: experiments.tex
\section{Experiments}{\label{sec:experiments}}
In this section, we experimentally compare Algorithm~\ref{algo:lev} with two variants of Algorithm~\ref{algo:lev}: $(i)$ the case where Step 4 is replaced with uniform/norm-based feature sampling and $(ii)$ the classical Breiman's RFs algorithm~\cite{breiman2001random}. Here, we highlight a subtle distinction between the RFs algorithm and all other algorithms under comparison. During the tree construction, RFs utilizes randomness for deciding the next split. Namely, at each node (assuming an additional split has been decided to be made), RFs selects $\lceil\sqrt{d}\rceil$ features uniformly at random on which the best split is selected. Therefore, one expects each resulting tree to possibly depend on all features as opposed to Algorithm~\ref{algo:lev} that depends on only $k$ features.
\vspace{0.05in}
%%%%%%%%%%%%%%%%%%%%%%%%
\begin{figure*}[!htp]
	\centering
	\begin{subfigure}{0.245\textwidth}
		\includegraphics[width=\textwidth]{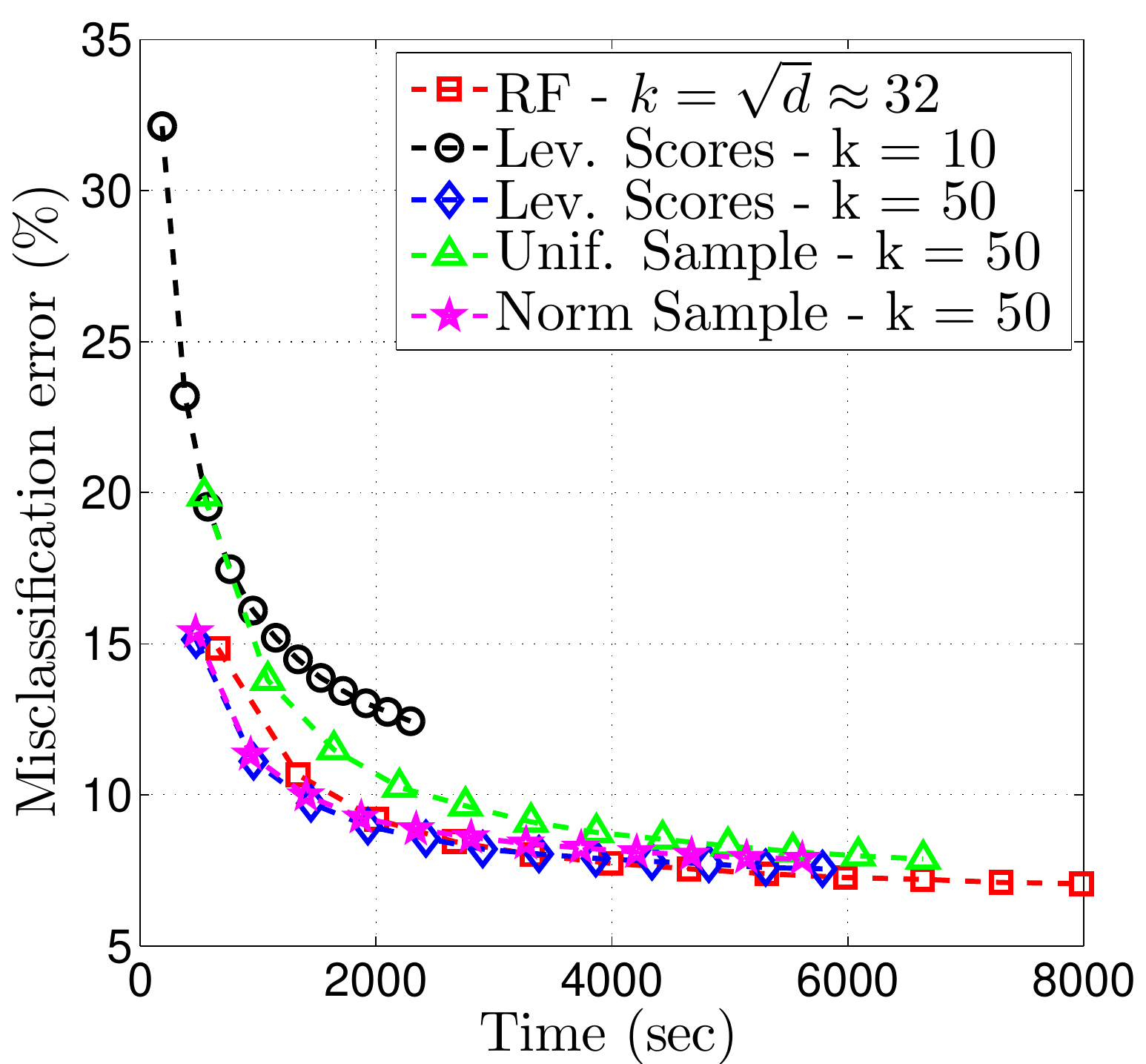}
	\end{subfigure} 
	\begin{subfigure}{0.245\textwidth}
		\includegraphics[width=\textwidth]{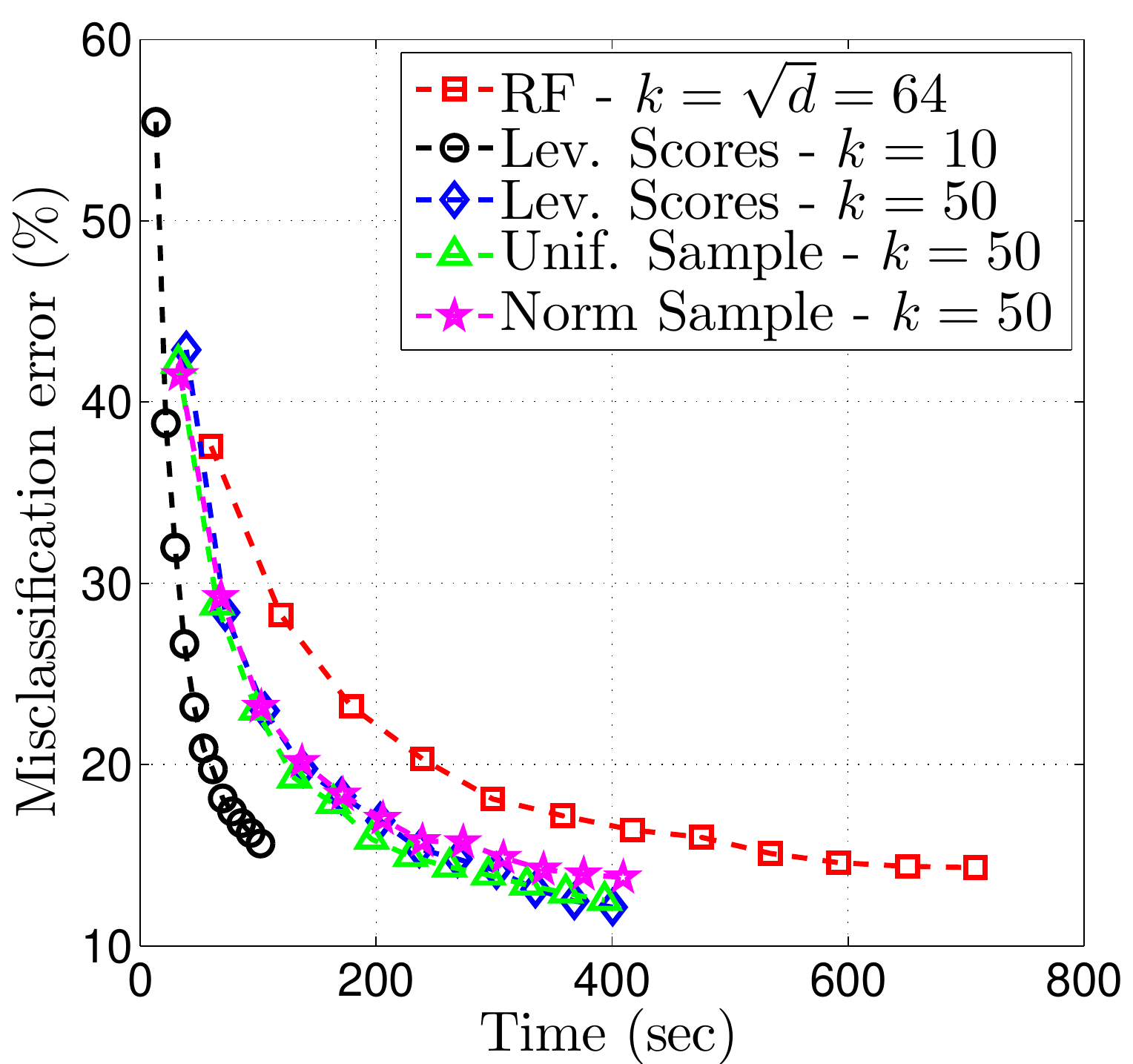}
	\end{subfigure} 
	\begin{subfigure}{0.245\textwidth}
		\includegraphics[width=\textwidth]{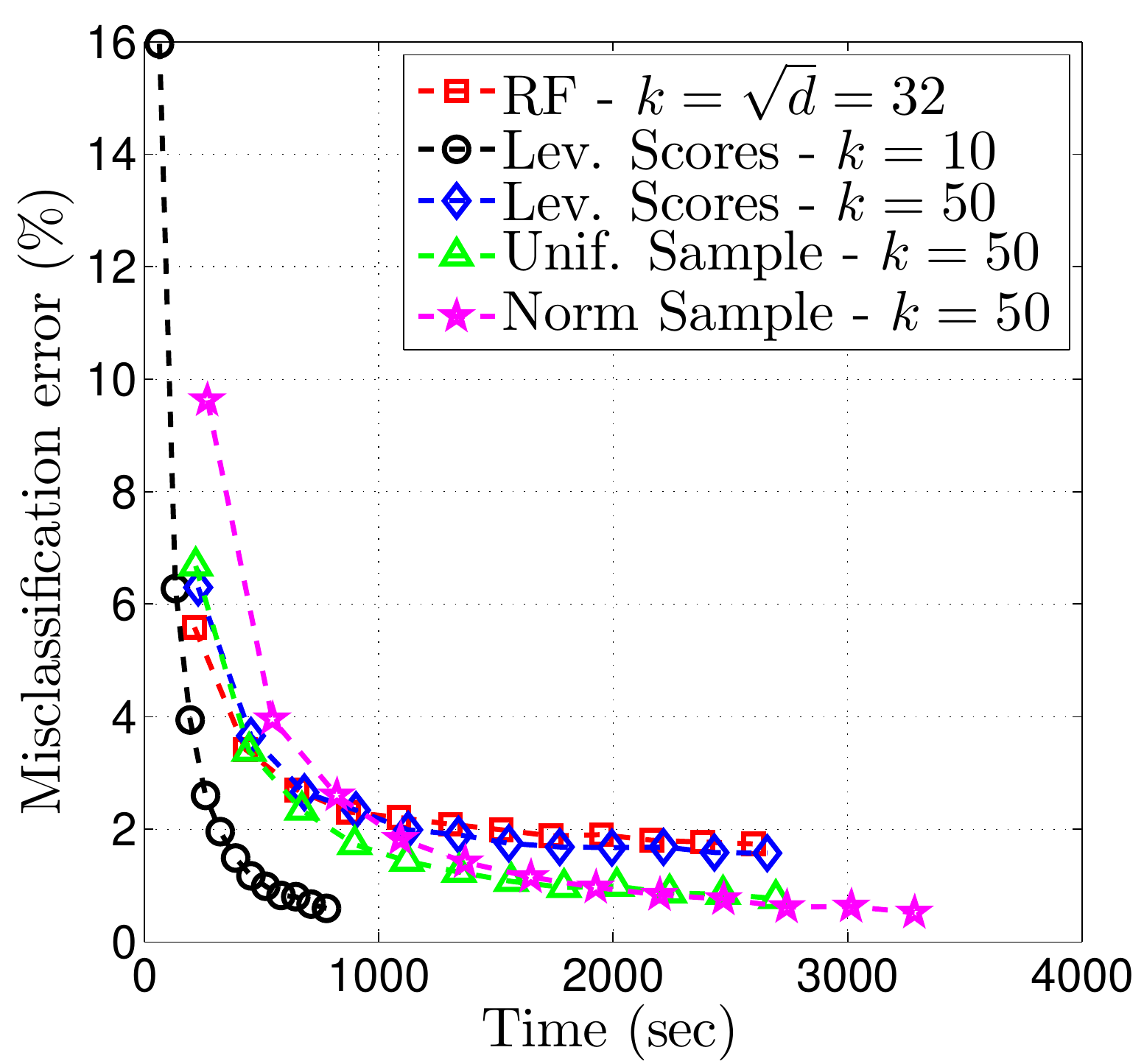}
	\end{subfigure} 
	\begin{subfigure}{0.245\textwidth}
		\includegraphics[width=\textwidth]{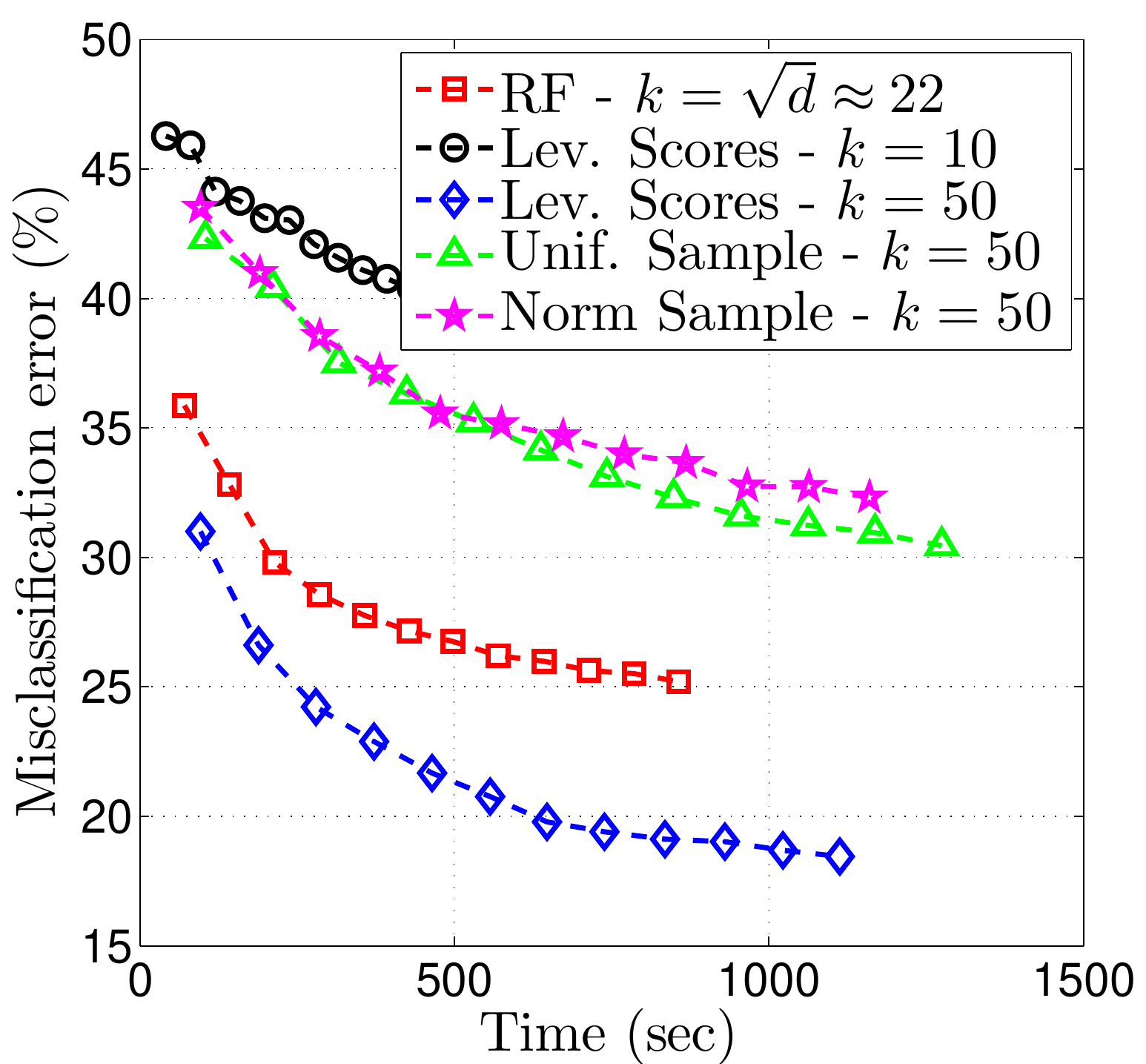}
	\end{subfigure} \vspace{.2in}\\
	\begin{subfigure}{0.245\textwidth}
		\includegraphics[width=\textwidth]{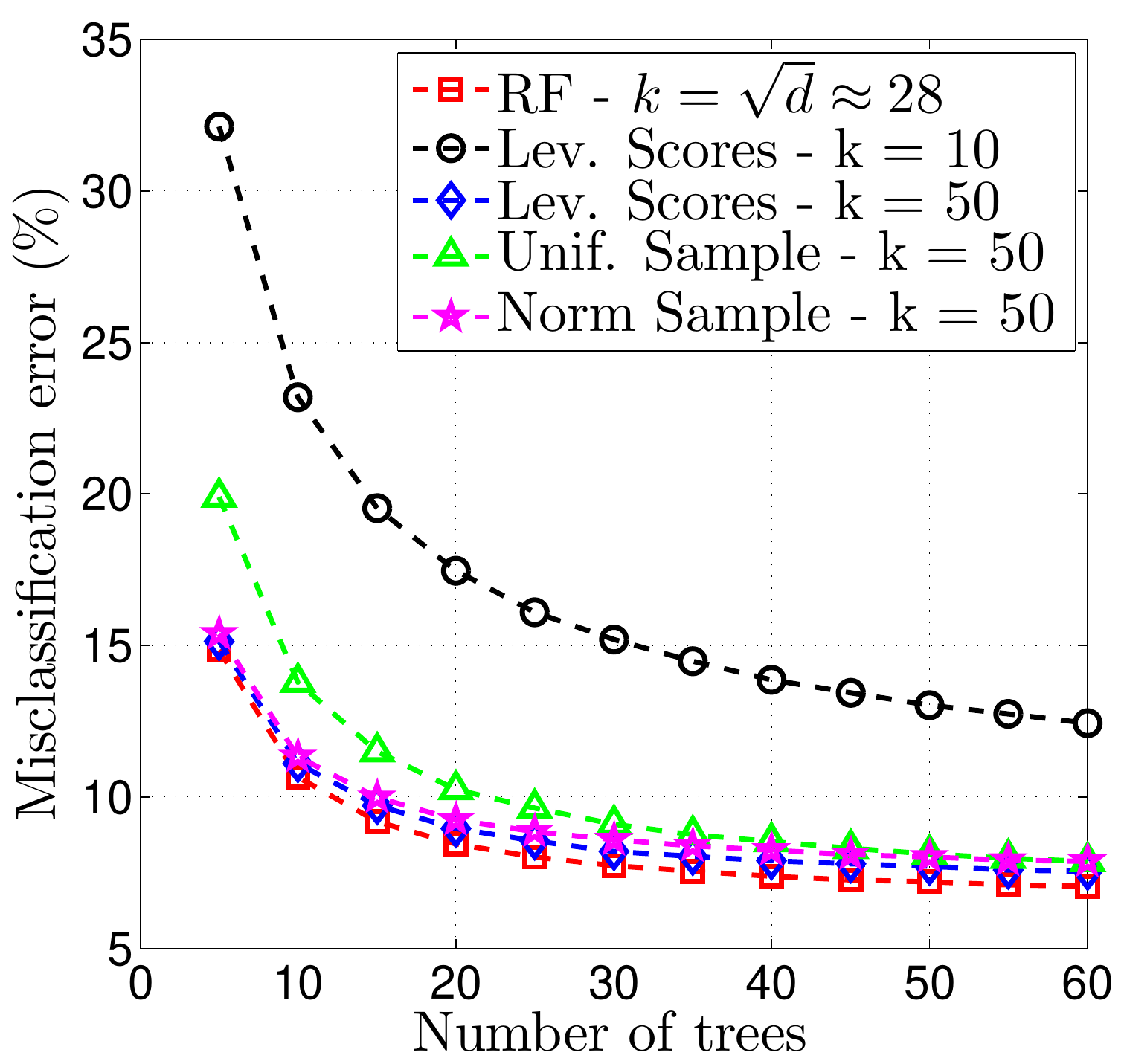}
		\caption{\small \texttt{MNIST}}
	\end{subfigure}
	\begin{subfigure}{0.245\textwidth}
		\includegraphics[width=\textwidth]{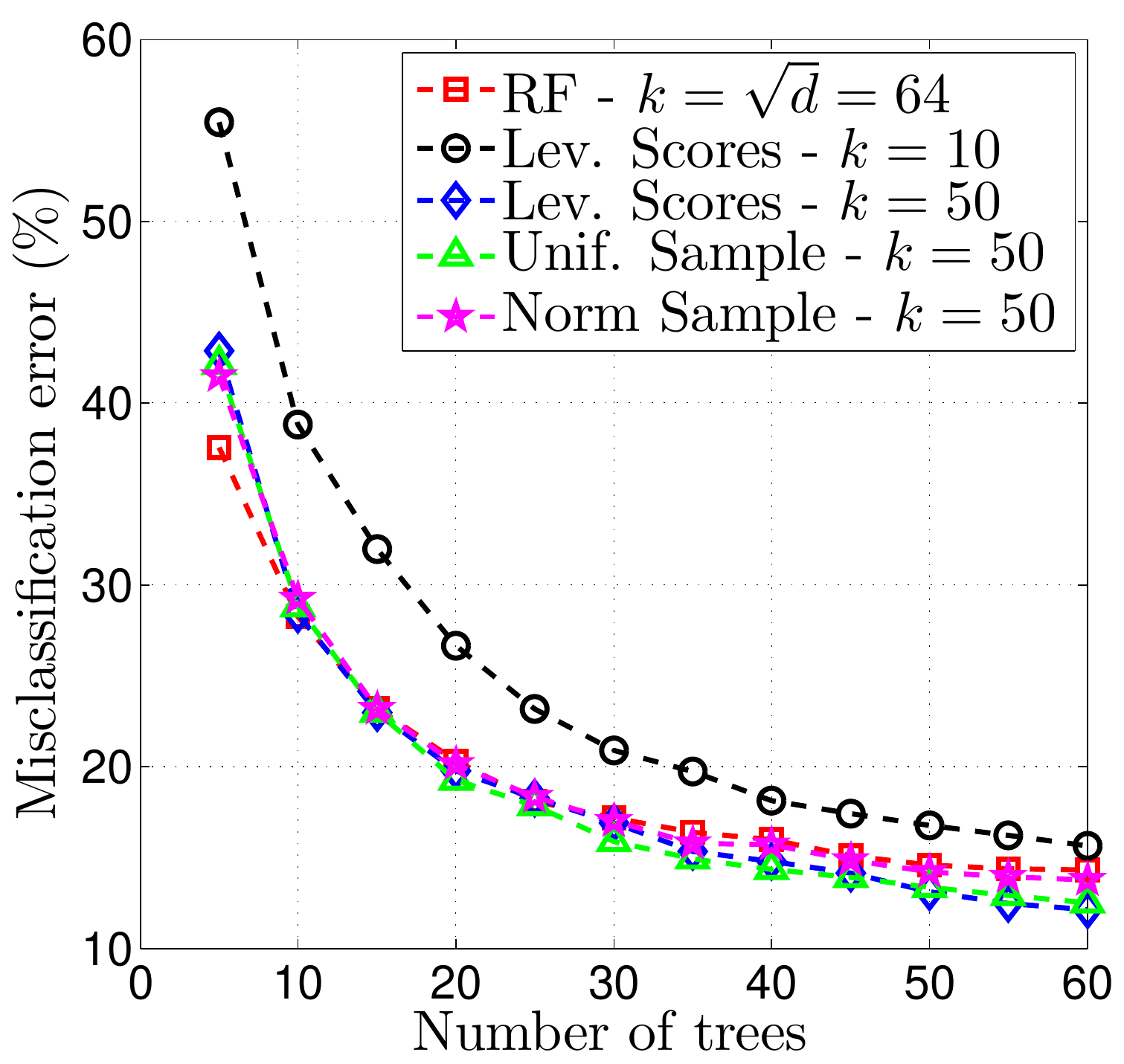}
		\caption{\small \texttt{ORL}}
	\end{subfigure}
	\begin{subfigure}{0.245\textwidth}
		\includegraphics[width=\textwidth]{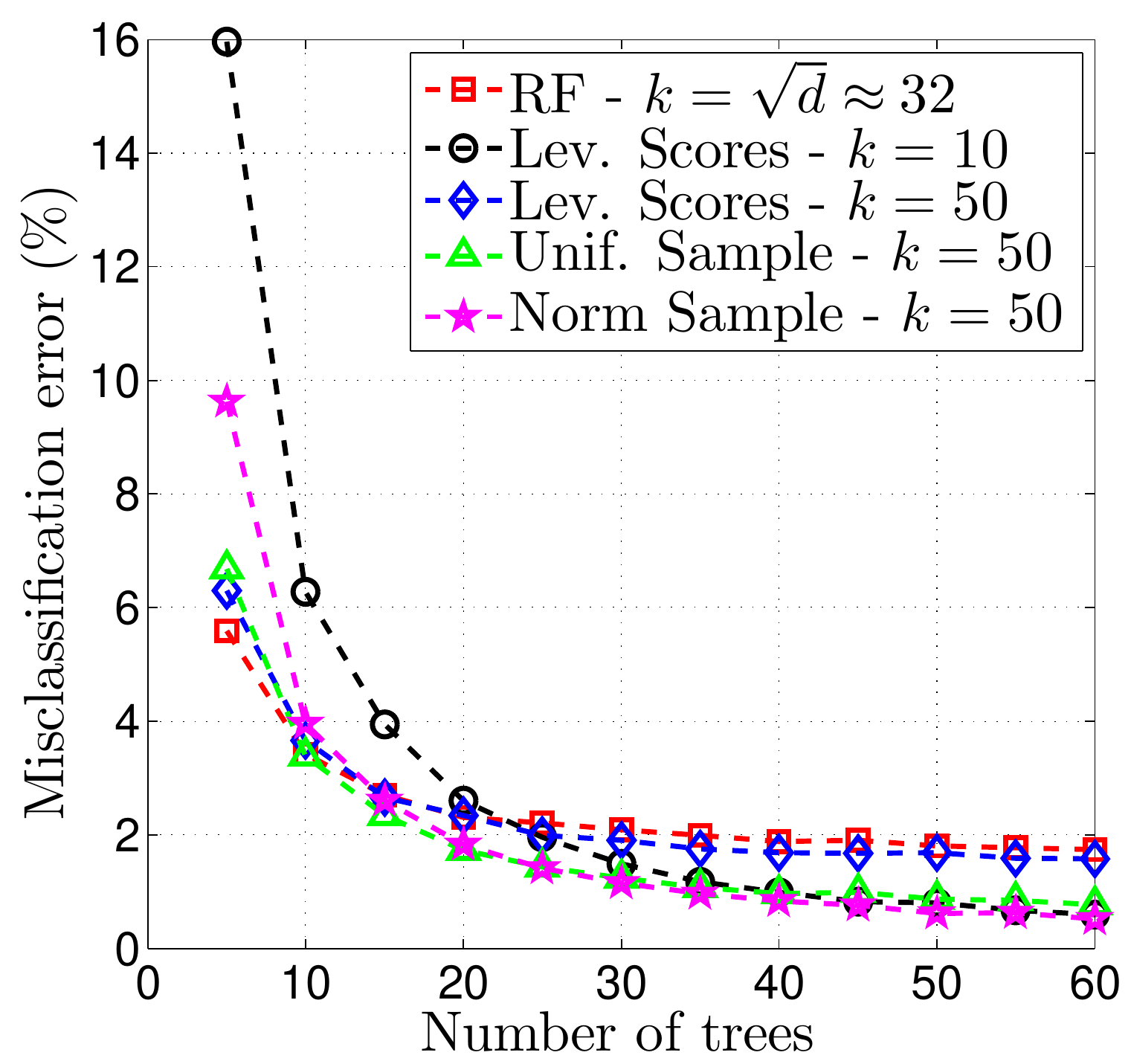}
		\caption{\small \texttt{PIE}}
	\end{subfigure}
	\begin{subfigure}{0.245\textwidth}
		\includegraphics[width=\textwidth]{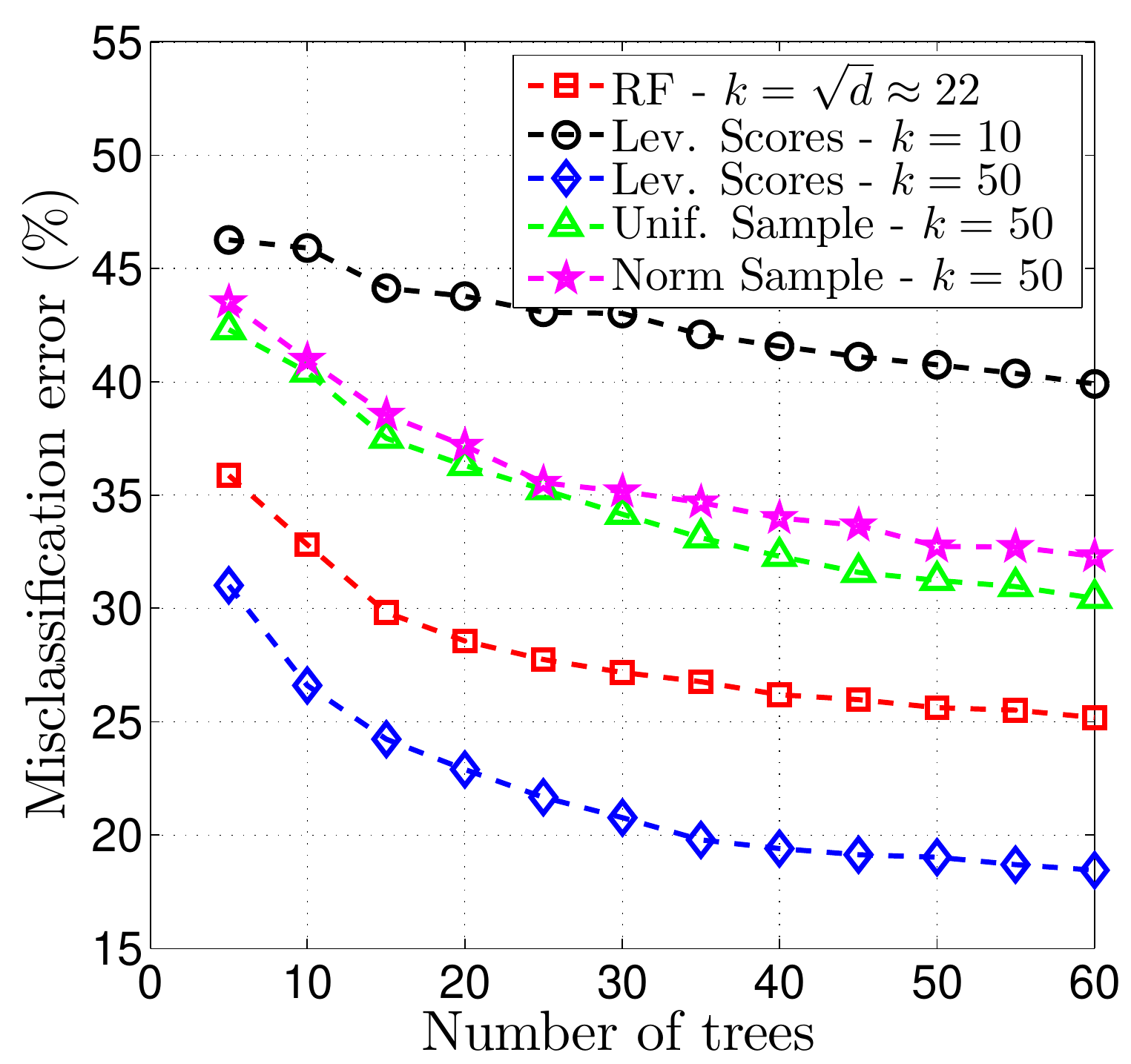}
		\caption{\small \texttt{MADELON}}
	\end{subfigure}
	\caption{\footnotesize Each column corresponds to a dataset. The first row depicts the classification error versus the elapsed training time for increasing number of trees. The second row depicts the classification accuracy versus the number of trees.}\label{fig:accVStime}
\end{figure*}
%%%%%%%%%%%%%%%%%%%%%%%%
%
\vskip.05in
\noindent \textbf{Datasets:} 
For the real-world datasets we used four publicly available\footnote{Most of the datasets used here are available under UCI's machine learning repository~\cite{UCI}.} datasets that we denote by \texttt{MNIST}, \texttt{ORL}, \texttt{PIE} and \texttt{MADELON}. The \texttt{MNIST} dataset of handwritten digits has a training set of 60,000 examples, and a test set of 10,000 examples~\cite{UCI} (a sample of $10,000$ training examples and $5,000$ test examples is used here). \texttt{ORL} contains ten different images each of $40$ distinct subjects~\cite{ORL}. There are 400 different objects in total, each having 4096 dimensions. \texttt{PIE} is a database of 41,368 images of 68 people~\cite{PIE}. Namely, there are in total $2856$ data points with $1024$ dimensions. The \texttt{MADELON} dataset is an artificial test case, multivariate and highly non-linear, part of the NIPS 2003 feature selection challenge. The data points of \texttt{MADELON} is $2000$ containing $500$ features. These datasets have been selected due to their high-dimensional feature space and/or their heavily usage as benchmarks for classification.
\vskip.05in
\noindent \textbf{Experimental setup:}
To measure the impact of leverage scores on the classification performance on random decision trees, we perform a series of diverse experiments on the above datasets. In all reported results, we use the average values of 30 independent executions of its corresponding algorithm. For the \texttt{LESS} algorithm, we truncate the computation of SVD to rank $r = 50$ for acceleration. We measured the performance of Algorithm~\ref{algo:lev} for various settings of $k$ and $t$, i.e., the number of features to be sampled and the number of trees, respectively. No bagging is performed on the RFs algorithm and the default number of subsampled features is selected, i.e., $\sqrt{d}$. All timings were performed under MATLAB R2011b~\cite{MATLAB:2011}.
%%
%\begin{itemize}
%\item [$(i)$] we set the number of training trees to $\lbrace 50, 100, 200 \rbrace$,\footnote{We observe empirically that, at least for the datasets that we consider in this writeup, the performance of RFs remains stable for $n > 200$ trees.}
%\item [$(ii)$] we modify the number of variables to be selected at each split node; we diverge from the rule-of-thumb heuristic where \texttt{NumVarToSample} $ = \sqrt{\text{\# \texttt{of features}}}$. Here, we consider cases where \texttt{NumVarToSample} = $\left\{ \left\lfloor\frac{\sqrt{\text{\# \texttt{of features}}}}{i}\right\rfloor ~|~ i=1,2,3,5 \right\}$.
%\item [$(iii)$] we modify the number of samples per node in the constructed random trees such that no further splitting is allowed, i.e., set the volume of the leafs. Using this feature, we implicitly determine the depth of the random trees and thus reduce/increase the space complexity of the learners. Here, the number of samples per node without further splitting is set to $\lbrace 100, 50, 20, 5, 1\rbrace$. For example, in the last case, we force the random decision trees to have a unique sample per leaf node.
%\end{itemize}
\vskip.05in
\noindent \textbf{Numerical results:} We report our experimental evaluations in Figure \ref{fig:accVStime} and in Table \ref{table:space}.
%($k = \lceil\sqrt{d}\rceil$) 
\begin{table}[!htp]
	\centering
	%\small
  \begin{tabular}{|c||c|c|c|c|c|}
	\hline
	\multicolumn{6}{ |c| }{Total number of nodes for given $\epsilon$ classification accuracy}    \\
    \hline
    \hline
	& $\epsilon$ & RFs & Uniform & Norm & Lev. Scores \\
    \hline
    \texttt{MNIST} & $\sim 7\%$  & 102066 & 125152 & \textcolor{red}{\textbf{92064}} & 93527 \\
	\hline
	\texttt{ORL} & $\sim 12\%$  & \textcolor{red}{\textbf{5498}} & 5568 & 5566 & 5587 \\
	\hline
	\texttt{MADELON} & $\sim 26\%$  & 26407 & $>$\texttt{T} & $>$\texttt{T} & \textcolor{red}{\textbf{4003}} \\
	\hline
	\texttt{PIE} & $\sim 1\%$ & \textcolor{red}{\textbf{18198}} & 20632 & 24851 & 19137 \\
	\hline
  \end{tabular}
\caption{\footnotesize Total number of nodes using $k = 50$ features.}\label{table:space}
\end{table}%
In the first row of Figure~\ref{fig:accVStime} we depict the classification error versus the elapsed training time for a pair of train and test data. The rationale behind this plot is to demonstrate that Algorithm~\ref{algo:lev} can achieve similar or better classification error than RFs with lower computational requirements. In all cases, \texttt{LESS} algorithm with $k=50$ is superior or matches the performance of RFs. On the other hand, the performance of Algorithm~\ref{algo:lev} with $k=10$ is not superior in all cases. This is due to the small value of $k$. Hence, a suggestive setting of $k$ in Algorithm~\ref{algo:lev} is in the range $[\sqrt{d},2\sqrt{d}]$. However, in stark contrast with conventional wisdom, there are cases where only $k = 10$ features, selected using leverage scores, seem to be sufficient to achieve the same or even better classification accuracy in much less training time, increasing the interpretability of the result due to the limited number of used features; e.g., see Figure \ref{fig:accVStime}(first row) for the cases \texttt{ORL} and \texttt{PIE}. Moreover, an increased number of features usually results in an increased processing time, with no further classification error improvement. Overall, we observe that \texttt{LESS} trees are at least as accurate as RF, while being less computationally expensive in practice. 

The second row of Figure~\ref{fig:accVStime} depicts the classification accuracy versus the number of trees. We observe that Algorithm~\ref{algo:lev} with $k=50$ matches the performance of RFs in terms of number of trees. Moreover, in the \texttt{MADELON} dataset Algorithm~\ref{algo:lev} is superior to RFs, which in turn, RFs is superior to both uniform and norm based feature selection. On the other hand, Algorithm~\ref{algo:lev} with $k=10$ does not perform well. 
%
%
%\begin{figure}[!h]
%	\centering
%	\includegraphics[width=0.25\textwidth]{figs/nodesVSfeatures_madelon_cropped}
%	\caption{\small Average number of nodes per tree versus the number of features for \texttt{MADELON}.}\label{rif:numTrees}
%\end{figure}
%

%
We further study the space complexity of the resulting ensembles as a function of the total number of nodes needed among all trees to achieve a predefined classification error $\epsilon > 0$. We also set up a time threshold limit value \texttt{T} $=3600$ seconds (1 hour) per approach to achieve accuracy $\epsilon$. Table~\ref{table:space} shows the reported space complexities for all test cases. As observed, both RFs and \texttt{LESS} trees has similar (or even better) space complexity for given $\epsilon$. From a different perspective, in a memory-limited scenario where only a fixed number of nodes can be maintained, non-uniform feature sampling leads to equivalent, if not better, mis-classification error level, as compared to uniform feature selection and/or RFs.

%% file: discussion.tex
\section{Discussion and Future Work}
In this work, we study feature selection strategies in classification, both in terms of time/space-complexity efficiency as well as of classification accuracy. Overall, results indicate that the proposed tree ensemble, based on leverage scores, might outperform the state-of-the-art RFs~\cite{breiman2001random}, as well as schemes where uniform weighting is applied. We observe that the proposed scheme results into low space-complexity trees for better interpretability, requires overall less training time and has at least the same accuracy, as compared to top-notch approaches.